\documentclass[letterpaper, 10 pt, conference]{ieeeconf}  

\usepackage{graphicx}
\usepackage{amsmath}
\usepackage{amssymb}
\usepackage{booktabs}
\usepackage{multirow}
\usepackage{subcaption}

\usepackage[capitalize]{cleveref}
\crefname{section}{Sec.}{Secs.}
\Crefname{section}{Section}{Sections}
\Crefname{table}{Table}{Tables}
\crefname{table}{Tab.}{Tabs.}

\IEEEoverridecommandlockouts                              

\title{\LARGE \bf
A Synthetic-Driven Vision System for Assembly Step Recognition
}

\author{
\authorblockN{Hui Zhang\textsuperscript{1,2,*,\ensuremath{\dagger}}, Xuanang Lei\textsuperscript{1,*}, Rui Wang\textsuperscript{1}, Julian Ferchow\textsuperscript{1,2}, and Mirko Meboldt\textsuperscript{1}}
\authorblockA{\textsuperscript{1}ETH Zurich, Switzerland; \textsuperscript{2}inspire AG, Switzerland}
\thanks{\textsuperscript{*}Equal Contribution.}
\thanks{\textsuperscript{\ensuremath{\dagger}}Correspondence email: huizhang@ethz.ch.}
}

\begin{document}

\maketitle
\thispagestyle{empty}
\pagestyle{empty}

\begin{abstract}

Quality control in industrial assembly is essential, and real-time monitoring of the assembly process is crucial for preventing costly defects and ensuring production reliability. Vision-based automated inspection offers a powerful solution for such real-time monitoring. However, due to the specialized industrial components and processes, training these models typically relies on task-specific real-world data, which is costly and labor-intensive to collect and annotate. In this paper, we propose a system that automatically generates realistic assembly sequences and further trains real-time inspection models using the synthetic data. It can be efficiently applied to a given task within an hour, requiring only CAD models and simple step descriptions. 
Focusing on practical challenges, our system integrates a physics-based motion generation module to capture the variance of different human assembly, designs domain-randomized rendering to deal with the environmental complexity and variation, and employs an object-detection-based step recognition module for robust sim-to-real transfer, leading to 92.4\% accuracy on a real-world assembly case with 46.7\%, 15.8\% and 61.2\% performance improvement, respectively.
Overall, our system provides a practical solution for industrial assembly inspection without requiring expensive real-world data collection and annotation, with the effectiveness validated on real industrial assembly tasks.
\end{abstract}


\section{Introduction}
\label{sec:intro}

High-mix low-volume production has become increasingly important in modern manufacturing, where manual assembly remains indispensable with its flexibility. However, manual assembly is inherently error-prone, as workers might miss steps, follow incorrect sequences, or misplace components, which can lead to costly defects if errors are not detected in real time.
Computer vision systems show promise for real-time assembly inspection \cite{Mazzetto_2020, s20154208}. Nevertheless, due to the specialized components and processes, training models for such systems requires task specific data, which is costly to collect and annotate in the real world, especially for high-mix low-volume production which requires fast adaptation. 

To address this limitation, leveraging synthetic data to train inspection models with minimal manual effort is highly desirable. However, several practical challenges remain in industrial assembly scenarios: (1) industrial components and assembly processes are highly task-specific, leading to large domain gaps from daily objects and across tasks, so general-purpose models often fail and task-specific data are required; (2) manual assembly involves continuous hand–tool–component interactions and severe occlusions, which are difficult to model and synthesize and have constrained prior works that only render static component combinations without hands, making them applicable only to simple tasks with fully visible parts and hand-free inspection \cite{Jonas, Zhu2024, rawal2024syntheticdatagenerationbridging}; (3) different workers adopt diverse strategies and factories exhibit frequent changes in lighting and backgrounds, demanding broad data coverage and model robustness to large motion and appearance variations; and (4) synthetic data inevitably suffers from sim-to-real gaps in texture, illumination, and motion realism, which become more pronounced in dynamic assembly processes within complex industrial environments and must be mitigated for practical deployment.

\begin{figure}
    \vspace{2mm}
    \centering
    \includegraphics[width=0.45\columnwidth]{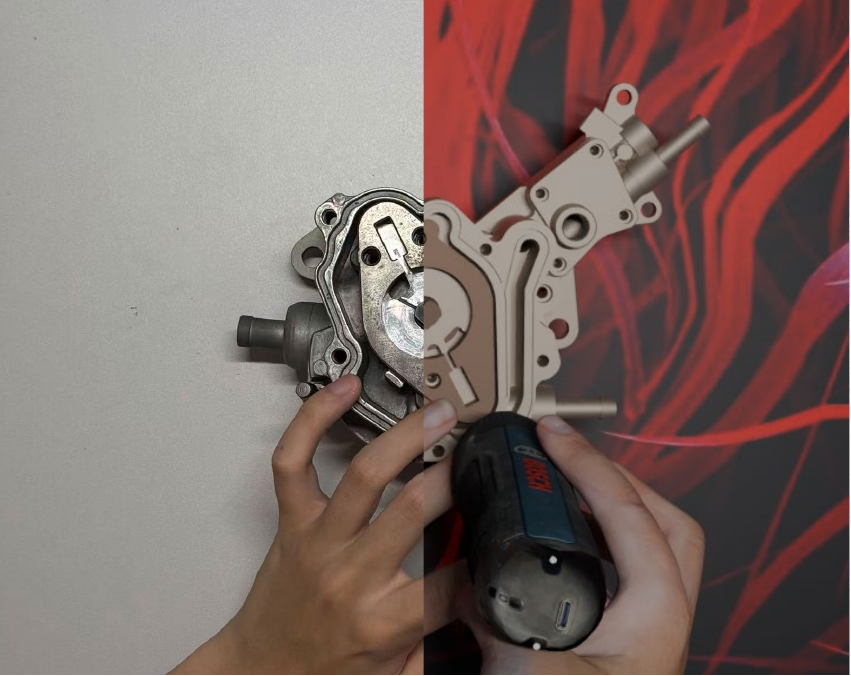}
    \caption{Real assembly data vs. Our synthetic data}
    \label{fig:teaser}
    \vspace{-6mm}
\end{figure}

To directly tackle these practical challenges, we propose a synthetic-data-driven system for industrial assembly step recognition, which is composed of three key modules: physics-based motion generation, photorealistic sequence rendering, and object-detection-based step recognition. 
Specifically, (1) to handle task specificity and domain gaps, we automatically construct task-specific assembly data for each step from the provided CAD models and step descriptions, which are used to train the inspection model; (2) to model complex, dynamic hand–tool–component interactions that prior static rendering pipelines cannot capture, we adapt the grasping motion generation method \cite{zhang2024graspxl} to produce diverse physics-based assembly motions in simulation; (3) to cover variability in worker strategies and environmental conditions, we generate diverse motions for each step and render them into photorealistic RGB sequences in Blender~\cite{blender} with randomized backgrounds, hand poses, and lighting; and (4) to mitigate sim-to-real gaps and achieve robust real-time deployment, we train a YOLO-based~\cite{yolov8_ultralytics} object detection model on the synthetic data and integrate it with a rule-based temporal filter for reliable step recognition in real videos.

We validate the proposed system through comprehensive experiments on a real industrial assembly task (vacuum pump). Since our goal is object-detection-based step recognition, we first evaluate the object detection performance of the trained model on both synthetic and real test sets to assess how well the system-trained detector transfers across the sim-to-real gap. We then evaluate the assembly step recognition performance of our system and compare it against alternative baselines and system variants, demonstrating clear advantages. We further analyze the robustness of the system across different human operators and under varying lighting conditions. These experiments confirm that the high-quality synthetic data and the integrated modules in our system jointly bridge the sim-to-real gap and enable reliable monitoring under practical variations in industrial assembly. Finally, we showcase the system’s generalizability by applying the full pipeline to a new assembly task (Framework laptop) with less than one hour of manual effort.

Overall, our contributions can be summarized as follows:
\begin{itemize}
    \item A complete system for assembly step recognition that, given only simple task specifications, efficiently synthesizes task-specific assembly data, trains detection models on the synthetic data, and generalizes to new assembly tasks with minimal manual effort
    \item A physics-based assembly motion generation module that captures dynamic hand-object interactions and human variances by simulating diverse assembly motions
    \item A photorealistic rendering module that produces high-quality RGB sequences with environmental variations 
    \item An object-detection-based assembly step recognition module which effectively deals with sim-to-real gaps and demonstrates robustness to human variances and environmental variations
\end{itemize}

\section{Related Work}
\label{sec:related_work}

\subsection{Industrial Assembly Inspection}
Industrial assembly inspection plays a critical role in quality and cost control, which has been a long-standing research problem \cite{509207, 5381107, 9383264}. With the advancement of computer vision, learning-based automatic assembly inspection with computer vision methods systems have been increasingly adopted to automatically recognize assembly states, detect errors, or verify task completion from visual observations \cite{Mazzetto_2020,s20154208,smirl}, reducing human labor and improving product quality. 
Despite promising performance, most existing systems rely on task-specific real-world data, requiring extensive data collection and fine-grained annotation for each new product or assembly process \cite{aganian2023attach}, which limits scalability and adaptability in practical industrial deployments. Furthermore, these systems often exhibit limited robustness to operator variability, viewpoint changes, and frequent hand-object occlusions inherent in manual assembly, posing challenges for reliable deployment on factory floors.

To reduce data collection costs, recent studies explore training inspection models using synthetic data \cite{Jonas,Zhu2024,rawal2024syntheticdatagenerationbridging,schieber2024asdf}. However, due to the difficulty of modeling the dynamic interactions between the hand, components, and tools, these works usually render static images with assembled components while ignoring dynamic assembly processes with complex hand-object interactions and occlusions, leading to a larger sim-to-real gap of the synthetic data. Consequently, such methods remain limited to simple assembly cases with clearly visible components and hand-free detection. 

In contrast, our work introduces a physics-based hand-object interaction synthesis module to generate dynamic assembly sequences, which are further rendered as training data for assembly step recognition.
By explicitly modeling contact, motion, and occlusion during assembly execution, the generated data narrows the sim-to-real gap and enables robust assembly step recognition in real-world assembly scenarios under occlusions and dynamic hand-object interactions, which effectively helps with the data scarcity problem in industrial assembly inspection.

\subsection{Hand-Object Interaction Recognition}
Hand–object interaction (HOI) recognition plays an important role in fields such as human–robot interaction and human activity understanding \cite{shamil2024handformer, FirstPersonAction_CVPR2018, Guilhem2015PCNN, wang2023pov}. Recent progress has been driven by the emergence of numerous HOI datasets \cite{fan2023arctic, liu2024taco, sener2022assembly101, fu2025gigahands, zhan2024oakink2}. However, action recognition in industrial assembly remains challenging, mainly due to the highly specialized components, tools, and processes involved. These factors introduce significant domain gaps not only between everyday HOI tasks and industrial assembly, but also across different assembly tasks themselves. As a result, training practical assembly inspection systems typically requires task-specific data with detailed annotations, and existing HOI datasets are not sufficient.

Synthetic data generation offers a promising alternative for creating large-scale, perfectly annotated training data, facilitating generalization to unseen scenarios. However, due to the difficulty of generating high-quality dynamic HOI data, current works in this field are still constrained to detecting static scenes of fixed objects without hands and interactions \cite{foundationposewen2024, wen2025foundationstereo}.
Manual assembly inspection, in contrast, involves dynamic hand-object interactions, which is more challenging to model and generate.
To address this, our work incorporates physics simulation to generate dynamic and temporally coherent assembly sequences, enabling step recognition during real-time dynamic assembly processes.

\begin{figure*}[t]
	\centering
        \vspace{2mm}
	\includegraphics[width=0.8\linewidth]{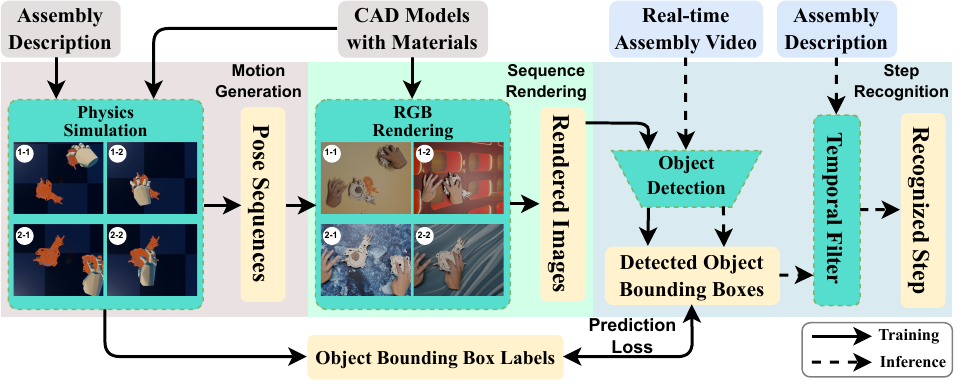}
	\caption{\textbf{Pipeline}. The pipeline consists of three modules: assembly motion generation, sequence rendering, and step recognition. During training, it takes assembly descriptions and CAD models with material properties as input, generates motions via physics simulation, renders RGB sequences in Blender \cite{blender}, and trains a YOLO-based object detector \cite{yolov8_ultralytics} with bounding box annotations. During testing, the trained model processes real-time assembly videos and produces detection results, which are mapped to step completion using a rule-based temporal filter guided by the assembly description. Importantly, the pipeline is not tailored to a specific task and can be efficiently adapted to new assembly cases.}
    \vspace{-4mm}
	\label{fig:pipeline}
\end{figure*}

\subsection{Hand-Object Interaction Synthesis}
The synthesis of realistic hand-object interactions (HOI) has gained significant attention in computer vision research due to its applications in AR/VR, robotics, and animation \cite{zhang2025bimart, zhang2025RobustDexGrasp, huang2024fungrasp}. Some works utilize data-driven methods to learn HOI synthesis based on collected datasets \cite{zhang2021manipnet, zheng2023cams}. Although enabling the generation of new HOI sequences, these methods usually suffer from physical inaccuracy problems such as penetration. More importantly, the generation capability is limited within the domain of the collected datasets, which constrains their application in real industrial assembly scenarios due to the significant domain gaps.

Instead of relying on expensive HOI datasets, recent advancements have enabled physics-based HOI synthesis with physics simulation and reinforcement learning \cite{zhang2024artigrasp, wang2025learning}, leading to better generalization across different object geometries and manipulation tasks. Specifically, the recent work GraspXL \cite{zhang2024graspxl} achieves remarkable generalization in generating grasping motions for 500k+ diverse objects with various poses, which builds up the foundation for generating more complex assembly motions with industrial components and tools. 
However, these works usually focus on generating 3D hand-object pose sequences and overlook the visual complexity of real-world environments (e.g., complex backgrounds and varying lighting), which limits their applicability for training vision-based inspection systems. In contrast, our work renders realistic motion sequences with practical factors such as occlusion, lighting, and background, producing synthetic datasets that more effectively bridge the sim-to-real gap for industrial assembly monitoring.

\section{Method}
\label{sec:method}

\subsection{Overview}

As illustrated in Figure~\ref{fig:pipeline}, our pipeline consists of three modules: physics-based assembly motion generation, photorealistic RGB sequence rendering, and object-detection-based step recognition, which are detailed in the following subsections.
During training, the pipeline takes two inputs: (1) CAD models of components and tools, including materials and textures, which inherently encode the final assembled states since they correspond to the completed product, and (2) a JSON-based assembly description specifying the sequence of steps together with the already assembled and to-be-assembled components and tools of each step, which can be manually created within 5 minutes. During inference, the trained model detects components and tools from real-time assembly video streams and identifies the current assembly step based on the detection results and assembly description.

\begin{figure*}
	\centering
        \vspace{2mm}
	\includegraphics[width=0.85\linewidth]{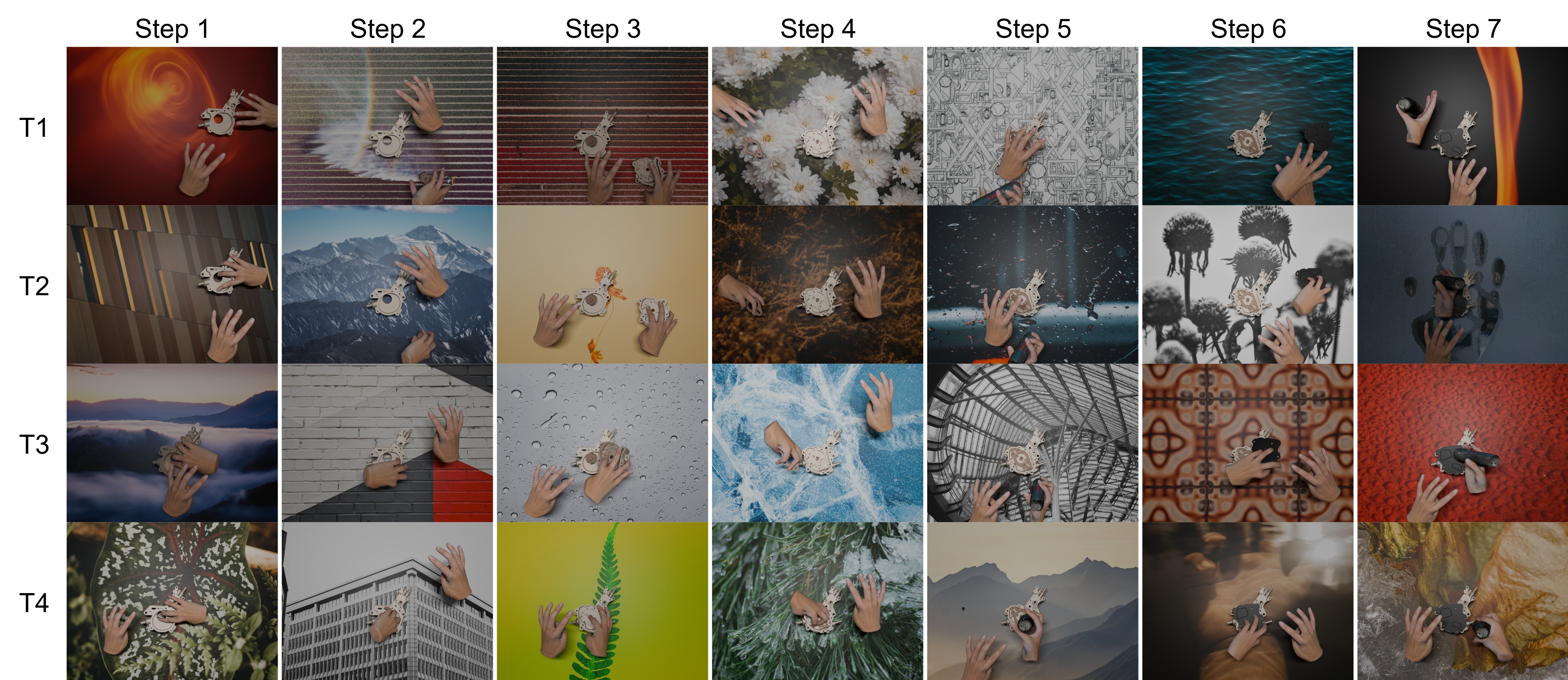}
	\caption{\textbf{Example of the rendered images}. We randomize the background and lighting conditions to simulate real-world variations. We also add random holding-hand occlusions, and apply motion blur and out-of-focus blur effects to simulate real camera artifacts. Each column represents an individual assembly step with four different time steps (T1-T4).}
    \vspace{-4mm}
	\label{fig:render}
\end{figure*}

\subsection{Assembly Motion Generation}
\label{sec:assembly_motion_generation}
In real-world assembly processes, the states of the hand, components, and tools evolve continuously over time, driven by dynamic physical interactions. To reduce the sim-to-real gap, it is crucial for synthetic data to capture these evolving state distributions and transitions, rather than relying on static snapshots or simplified object placements. However, accurately modeling realistic hand–object interactions remains challenging due to the complexity of contact dynamics, as well as the diversity in grasping strategies and tool usage patterns across different operators.

To address this, we build on GraspXL \cite{zhang2024graspxl}, a recent physics-based grasp motion synthesis method, and extend it to assembly motion synthesis. Specifically, for each assembly step, we fix the already assembled components specified in the assembly description on the table, and place the to-be-assembled component as a free object at a randomly sampled location on the table. 
GraspXL then controls a left or right hand to grasp the to-be-assembled component and, applying PD control at the wrist, transport it towards the target configuration specified by the CAD model, where it is integrated with the fixed assembled components. 
Since the entire process is simulated in a physics engine with realistic contact, friction, and gravity, all state transitions are inherently force-driven, resulting in physically realistic hand–object interactions. 
We represent the hand using the MANO model \cite{MANO:SIGGRAPHASIA:2017} with the mean shape, a standard choice in human motion synthesis.

This physics-based assembly motion generation is critical for producing realistic and diverse synthetic training data, particularly in modeling fine-grained spatial relationships, continuous pose changes, and operator-specific motion variations. Its effectiveness is validated in our ablation study (Section~\ref{sec:step_recongnition}). The generated sequences are stored as temporally continuous hand–object pose trajectories and serve as input to the subsequent rendering module for annotated frame generation. We demonstrate some of the generated motions in our supplementary video.

\subsection{Assembly Sequence Rendering}
With the generated motions, we further render the training images for the model. The rendering process imports component and tool CAD models and motion sequences generated from the previous module, and outputs photorealistic RGB assembly images. We first generate the 3D hand meshes of each frame according to the recorded hand poses and MANO model \cite{MANO:SIGGRAPHASIA:2017}. Then, we use Blender \cite{blender} to render the images, according to component/tool/hand materials and textures defined in the CAD models and the rendering parameters including lighting, background, etc. 

To enhance the coverage of the synthetic data, which is essential to improve the robustness of the trained model, we implement domain randomization to the rendering process. Specifically, we randomize the lighting conditions using HDRI environment maps.
Besides, as the assembly can be performed at different locations with different backgrounds, and random components can sometimes show on the tabletop, we diversify the background with random images so that the model can largely discard background variations and instead focus on the assembly components, hands, and tools.
We also introduce motion blur and out-of-focus blur effects to simulate real camera artifacts. Besides, we add the hand used to hold the already assembled components at random positions around the component to simulate the occlusion.

Some examples of the rendered images are shown in Figure \ref{fig:render}, with more demonstrated in the supplementary video. As we render the images from physics-based assembly motions, it has a good coverage of the real-world hand-component and hand-tool spatial relationship distributions during assembly scenarios. This is crucial for the model to robustly recognize the assembly steps, which is verified by the experiments in Section \ref{sec:step_recongnition}.

In addition to image generation, our pipeline automatically produces step labels and oriented bounding box annotations for all components and tools in each frame. This automated labeling eliminates the need for manual annotation, which is crucial for scaling up to large datasets.

\subsection{Assembly Step Recognition}
To achieve model robustness against human variance and environmental diversity, our approach generates diverse assembly motions through physics-based simulation and renders them with extensive domain randomization as described in the previous sections. This synthetic data comprehensively covers various assembly scenarios, providing a rich training dataset.
Using the synthetic data, we train a single object detection model based on YOLOv8n \cite{yolov8_ultralytics}, which performs real-time detection of oriented bounding boxes for the different components and tools in each assembly step.

To enable robust step completion recognition based on object detection results, we implement a rule-based temporal filter that determines whether an assembly step is complete. Empirically, a step is considered complete when the required components and tools are consistently detected, with confidence $>$ $\alpha$, and positioned correctly, with the IoU $>$ $\beta$ between the bounding boxes of the assembled and to-be-assembled components. $\alpha$ and $\beta$ are two parameters that can be adjusted to optimize performance for specific industrial assembly scenarios.
To suppress detection noise and filter out false positives, we apply a temporal consistency check: a step is marked as complete only if the relevant objects remain in the correct position for at least 8 out of the past 10 frames. This temporal filtering helps improve the system's robustness in real-world industrial settings, as demonstrated by the experiments in Section \ref{sec:step_recongnition}.

\section{Experiment}
\label{sec:experiment}

\subsection{Experimental Setup}
\subsubsection{Assembly Environment}
\begin{figure}
    \centering
        \vspace{2mm}
    \includegraphics[width=0.45\columnwidth]{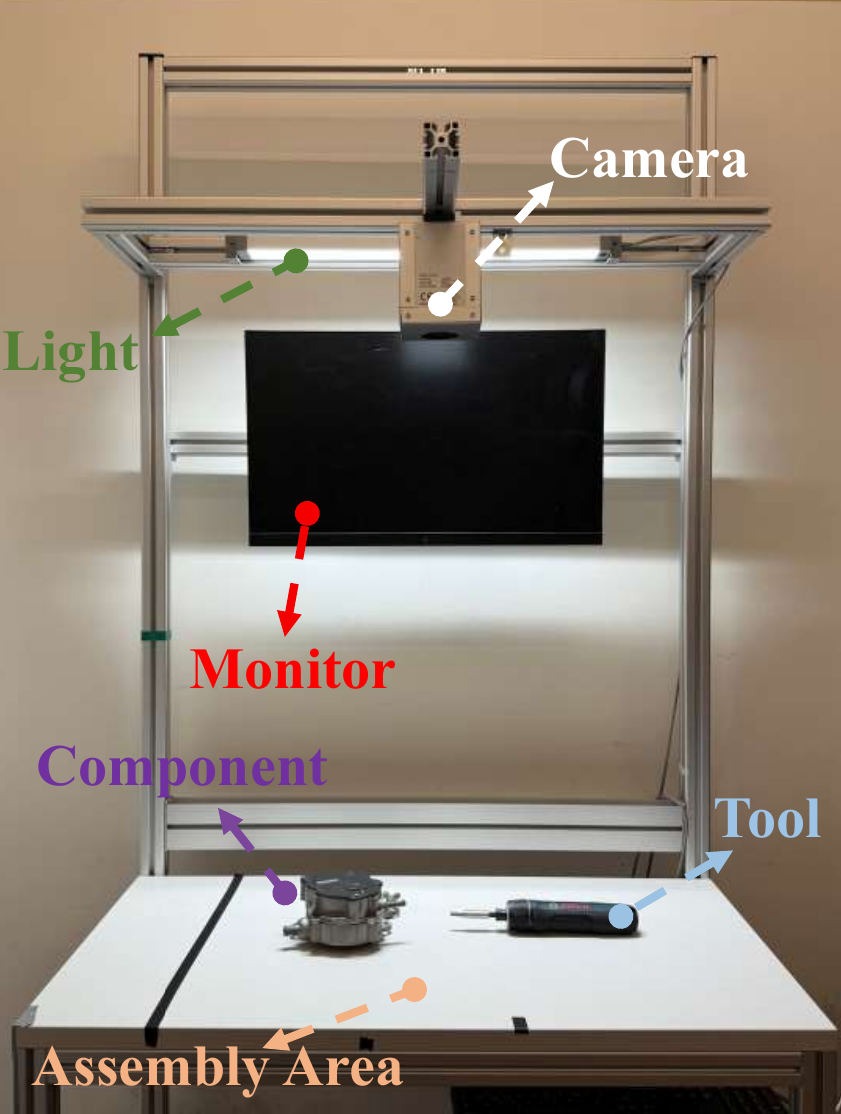}
    \caption{The assembly environment.}
    \vspace{-5.5mm}
    \label{fig:workbench}
\end{figure}

To validate our approach, we conduct experiments using a standardized industrial assembly workbench, which is illustrated in Figure~\ref{fig:workbench}. The workbench features a top-view RGB camera for real-time video stream recording, a light for illumination, a monitor for the user interface, and an assembly area as the workspace.

\subsubsection{Assembly Test Case}
We evaluated our system mainly on a representative industrial assembly task: a vacuum pump, which serves as a challenging test case due to its fully metallic composition, leading to strong specular reflections and visually similar components. As illustrated in Figure~\ref{fig:recognition_vacuum_pump}, the assembly process consists of seven sequential steps:
(1) Place Bottom Case;
(2) Assemble Axle Part;
(3) Assemble Upper Part;
(4) Place Diamond Component;
(5) Tighten Screws for Upper Part;
(6) Place Cover Case;
(7) Tighten Screws for Cover Case.
Note that we detect the completion of each assembly step, and do not track the exact number of screws tightened during the screw-fastening steps.

To ensure comprehensive coverage of the test data, we collect 24 assembly sequences from 6 different operators with diverse demographics in terms of gender, age, assembly experience, and handedness, who have no prior knowledge with this assembly task. We further manually annotate each image with step labels and object detection labels (without bounding boxes), resulting in a total of 53,569 annotated images for testing. Each operator performs the assembly task four times: twice under normal lighting conditions (light on) and twice under dim lighting conditions (light off).

\subsubsection{Implementation Details}

Assembly motions are generated in a physics simulator RaiSim~\cite{Hwangbo2018Raisim}. For each assembly step, we generate 30 diverse motion sequences, each with 130 time frames. Based on this, we render 3900 diverse images with randomized lighting conditions and backgrounds for each assembly step, leading to a total of 27300 various synthetic images. We randomly sample 70\% of the synthetic images to train the YOLOv8n-based \cite{yolov8_ultralytics} object detection model, leading to a train set of 19110 images. Notably, the model runs at over 200 fps on a single NVIDIA RTX 4090 GPU during inference, enabling real-time inspection.

\subsection{Object Detection Evaluation}
We first evaluate the performance of our object detection model on both real and synthetic test sets. For the synthetic test set, we use the remaining 30\% images not used for training with the auto-generated object bounding boxes. For the real test set, we use the recorded 53,569 images with the object detection labels (without bounding boxes) to evaluate the performance of our object detection model.

\subsubsection{Metrics}
We evaluate the performance with the following metrics: Precision ($\frac{\text{TP}}{\text{TP} + \text{FP}}$), Recall ($\frac{\text{TP}}{\text{TP} + \text{FN}}$), and mean Average Precision (mAP, following COCO protocol across IoU thresholds 0.5-0.95, only for the synthetic test set).

\subsubsection{Results}

\begin{table}[t]
    \centering
        \vspace{2mm}
    \resizebox{0.99\columnwidth}{!}{
    \begin{tabular}{l|ccc}
        \toprule
        \textbf{Test Set} & \textbf{Precision (\%)} & \textbf{Recall (\%)} & \textbf{mAP (\%)} \\
        \midrule
        Synthetic Test Set & 99.46 & 98.54 & 95.28  \\
        Real Test Set & 97.38 & 75.49 & - \\
        \bottomrule
    \end{tabular}
    }
    \caption{Object detection evaluation}
    \vspace{-3mm}
    \label{tab:object_detection_results}
  \end{table}

The results are presented in Table~\ref{tab:object_detection_results}. Our method achieves high performance on synthetic data, with 99.46\% precision, 98.54\% recall, and 95.28\% mAP, demonstrating accurate object identification and localization. Notably, despite being trained solely on synthetic data, the model generalizes well to real data, achieving 97.38\% precision and 75.49\% recall, indicating successful sim-to-real transfer. In particular, the comparable precision (97.38\% vs. 99.46\%) highlights the effectiveness of our method to close the sim-to-real gap.
The relatively lower recall on real data (75.49\% vs. 98.54\%) suggests that some objects in real-world scenarios are more challenging to detect, likely due to factors such as serious specular reflection caused by the metallic surfaces of the vacuum pump components, which are hard to replicate accurately in rendering.

\begin{figure*}[!th]
	\centering
\vspace{2mm}
    \begin{subfigure}[b]{0.78\linewidth}
		\centering
		\begin{tabular}{c}
			\includegraphics[width=0.92\linewidth]{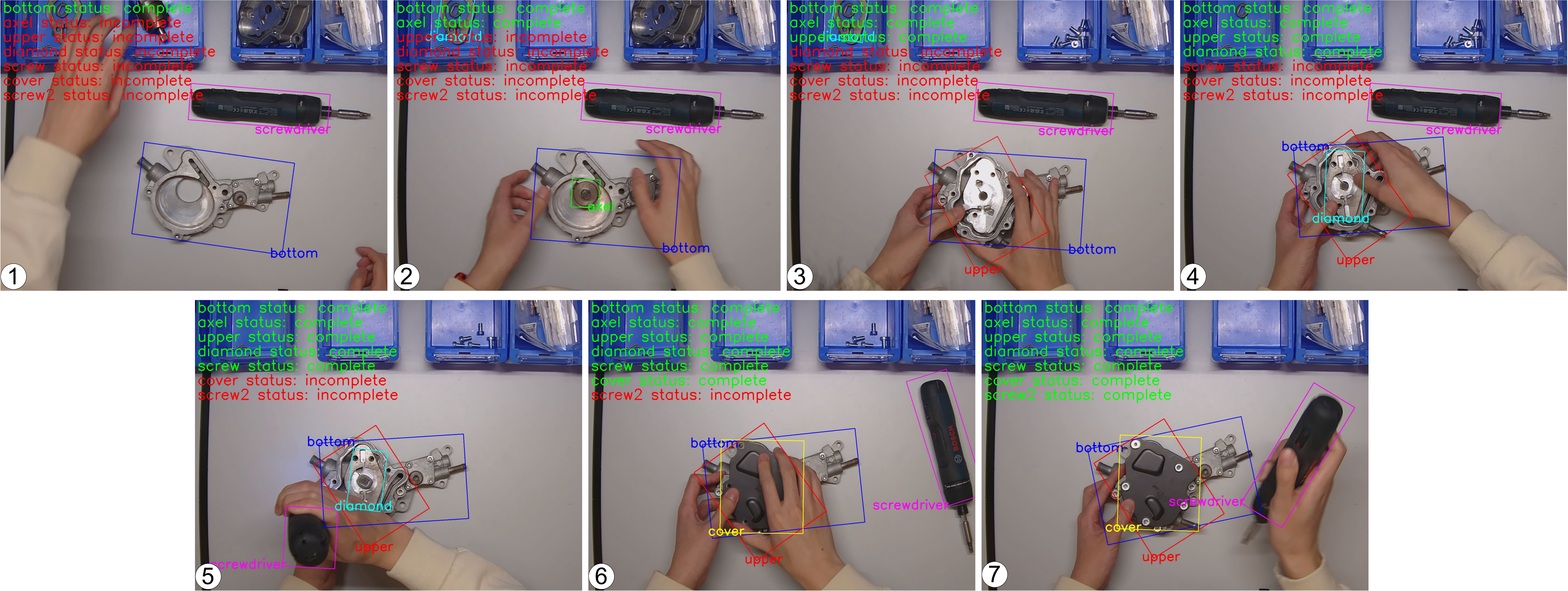}
		\end{tabular}
		\vspace{-3mm}
        \caption{Real-time object detection and step recognition for the vacuum pump.}
		\vspace{0.5mm}
		\label{fig:recognition_vacuum_pump}
	\end{subfigure}
	\begin{subfigure}[b]{0.94\linewidth}
		\centering
		\begin{tabular}{c}
			\includegraphics[width=0.94\linewidth]{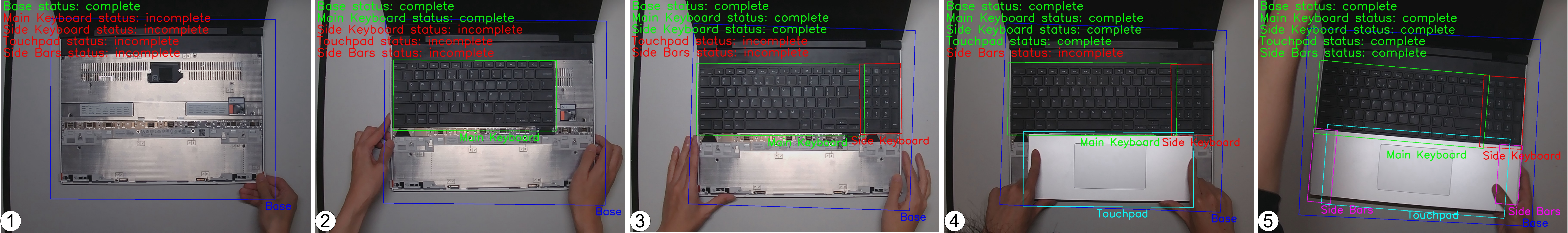}
		\end{tabular}
		\vspace{-3mm}
        \caption{Real-time object detection and step recognition for the Framework laptop.}
		\label{fig:recognition_framework}
	\end{subfigure}
    \vspace{-1mm}

	\caption{The system continuously tracks object placements and tool usages, and further determines step completion.
    }
	    \vspace{-4mm}
	\label{fig:recognition}
\end{figure*}

\subsection{Step Recognition Evaluation}
\label{sec:step_recongnition}
In this section, we evaluate the step recognition performance of our pipeline based on the object detection results and rule-based temporal filter, using our recorded 53,569 real-world images with step labels.
We use accuracy as the metric, which is the ratio of correctly recognized steps to total steps ($\text{Accuracy} = \text{Correctly Identified Steps}/\text{Total Steps}$).

\subsubsection{Method Comparison}

To validate the contribution of the components in our method, we first ablate the key components in our three modules and evaluate their performance:

\textbf{Object Detection without Motion Generation (Detect.(W.o. Motion)):} We ablate the motion generation module by directly rendering the target state of each step without simulated continuous hands and object trajectories. All other settings (e.g., rendering, temporal filtering) are unchanged.

\textbf{Object Detection without Random Lighting Rendering (Detect.(W.o. Randomization)):} We ablate lighting randomization in the rendering module by generating images under fixed lighting conditions, while keeping all other settings (e.g., motion generation, temporal filtering) unchanged. 

\textbf{Object Detection without Temporal Check (Detect.(W.o. Temporal)):} We ablate temporal consistency check by marking a step as complete once relevant objects are detected in the correct positions, without requiring multi-frame consistency. All other settings (e.g., motion generation, rendering, training) are unchanged.

To further verify the effectiveness of our object-detection-based step recognition, we compare it with end-to-end YOLO-based step classification baselines, which directly predict the current assembly step from input images without explicitly modeling object detection or spatial relationships:

\textbf{Classification with Real Data (Class.(Real)):} This baseline trains a classification model on 19,956 annotated real assembly images from 10 sequences (5 under normal lighting and 5 under dim lighting) performed by extra operators not included in the test set. Similar to our method, a step is considered complete only if the predicted label is correct for at least 8 out of the past 10 frames.

\textbf{Classification with Synthetic Data (Class.(Syn.)):} Similar to the previous baseline, this baseline utilizes a classification model but is trained exclusively on our motion-based synthetic images, supervised by the current step labels. The temporal consistency verification is the same.

\subsubsection{Results}
We provide the qualitative results in Figure~\ref{fig:recognition_vacuum_pump} and quantitative results in Table~\ref{tab:step_recognition_results}. More qualitative demonstrations can be found in our supplementary video.
Compared to the variant without motion generation in simulation (Detect.(W.o. Motion): 65.7\%), our method achieves a substantial improvement of 26.7\%. This clearly demonstrates the benefit of incorporating physics-based motion into synthetic data generation. Specifically, simulated motions effectively capture the changing spatial relationships between the hand, components, and tools during assembly, especially the frequent partial occlusions between them. Instead of rendering objects only at random, fully visible poses, we generate training images along realistic motion trajectories, which better match real assembly where parts and tools are often partially covered by the hand or other components. This makes the detector significantly more robust to severe occlusions and leads to higher accuracy for occluded tools and parts. The benefit is particularly pronounced in our challenging case, where many components are small and share very similar colors; in these scenarios, motion-based training reduces both false positives and missed detections, and allows the model to implicitly infer the occluded regions (e.g., reasoning about the shape and pose of a grasped part from the pose of the hand). Together, these effects help further narrow the sim-to-real gap. 

The model trained without random lighting (Detect.(W.o. Randomization): 76.6\%) shows a performance drop of 15.8\%, particularly on test sequences captured under dim lighting. This highlights the importance of domain randomization in our rendering pipeline, which enables the trained model to generalize robustly to diverse real-world conditions.
Additionally, removing the temporal consistency check (Detect.(W.o. Temporal): 75.0\%) results in a performance drop of 17.4\%, emphasizing the importance of temporal verification. Without it, the system becomes more susceptible to transient false positives in object detection, undermining the reliability of step completion recognition.

\begin{table}[t]
    \vspace{2mm}
    \centering
    \resizebox{0.75\columnwidth}{!}{
    \begin{tabular}{l|c}
        \toprule
        \textbf{Method} & \textbf{Accuracy (\%)} \\
        \midrule
        Detect.(W.o. Motion) & 65.7 \\
        Detect.(W.o. Randomization) & 76.6 \\
        Detect.(W.o. Temporal) & 75.0 \\
        \midrule
        Class.(Real) & 78.4 \\
        Class.(Syn.) & 31.2 \\
        \midrule
        Ours & 92.4 \\
        \bottomrule
    \end{tabular}
    }
    \caption{Step recognition evaluation}
    \vspace{-6mm}
    \label{tab:step_recognition_results}
  \end{table}

Classification models, even when trained on real data (Class.(Real): 78.4\%), underperform compared to our object-detection-based method (92.4\%). With synthetic training (Class.(Syn.): 31.2\%), their accuracy drops sharply, reflecting vulnerability to the sim-to-real gap.
This is primarily because classification models rely on global image features, which are inherently sensitive to variations in lighting, cluttered backgrounds, hand poses, and other environmental factors. Robustness of classification, therefore, typically requires large and diverse datasets that cover a wide range of visual variations, leading to lower accuracy even with real training data of the same size as our synthetic training set. Moreover, these global features are particularly hard to synthesize faithfully, making classification models more susceptible to the sim-to-real gap when trained on synthetic data.
In contrast, our detection-based method explicitly focuses on the task-relevant components and tools, filtering out irrelevant scene variations and generalizing more robustly across domains. Overall, this targeted focus helps reduce the reliance on large-scale real data and narrows the sim-to-real gap.
Furthermore, some assembly steps have almost identical visual appearances (e.g., before and after screw tightening), which makes them difficult for frame‑based classifiers to distinguish. Instead of relying on global visual features, our method detects the relevant components and tools and tracks their state changes over time, and then determines whether the current step is completed from these temporal object‑level transitions, which is not affected by the similar visual frames of different steps.

\subsection{Robustness Evaluation}
To evaluate the robustness of our method, we report and analyze the performance of our method among the different human operators and lighting conditions in the real test set.

\subsubsection{Robustness to Human Variance}

\begin{figure}[t]
	\centering
\vspace{2mm}
\begin{subfigure}[b]{0.99\linewidth}
    \centering
    \includegraphics[width=0.31\linewidth]{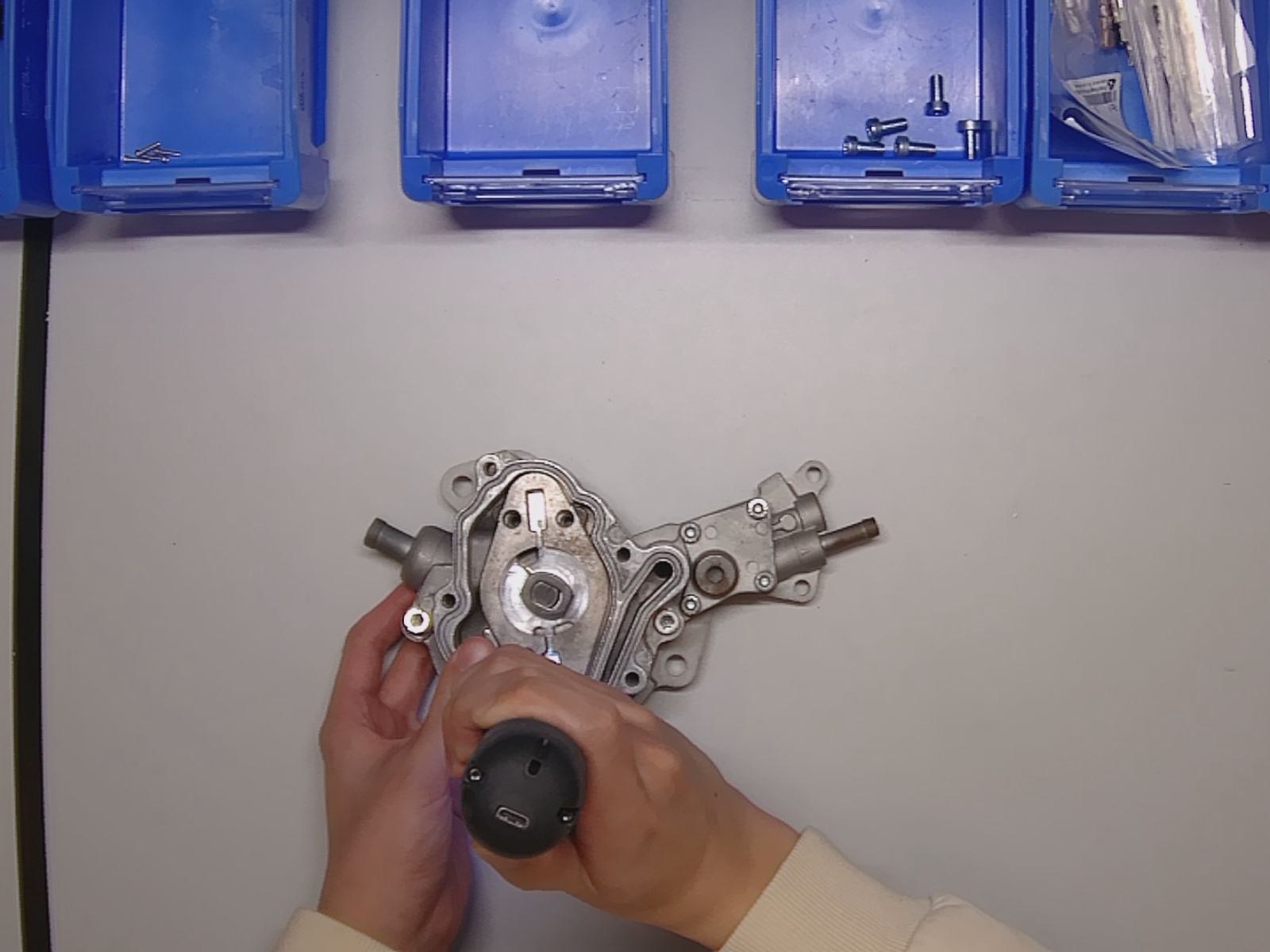}
    \includegraphics[width=0.31\linewidth]{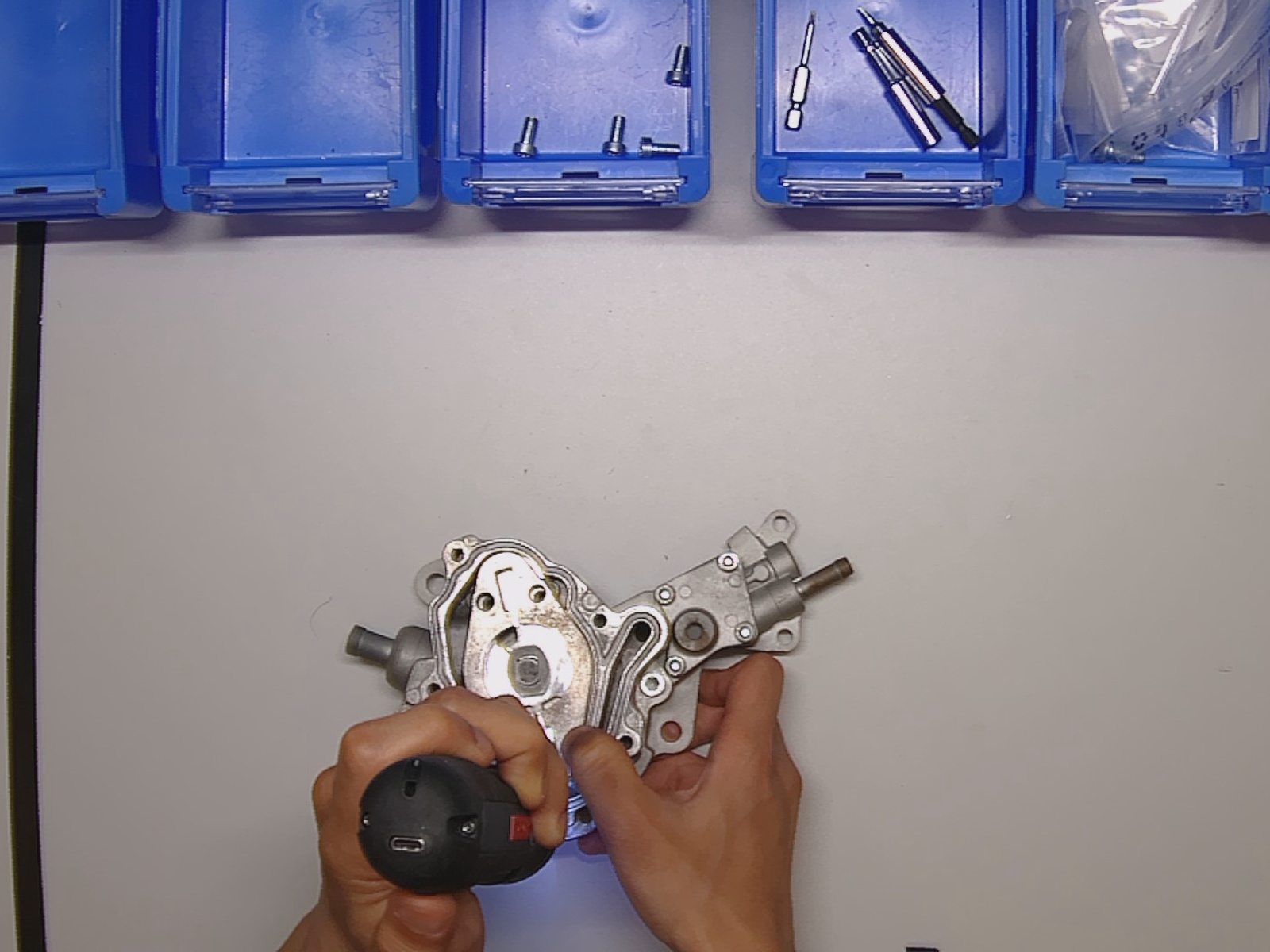}
    \includegraphics[width=0.31\linewidth]{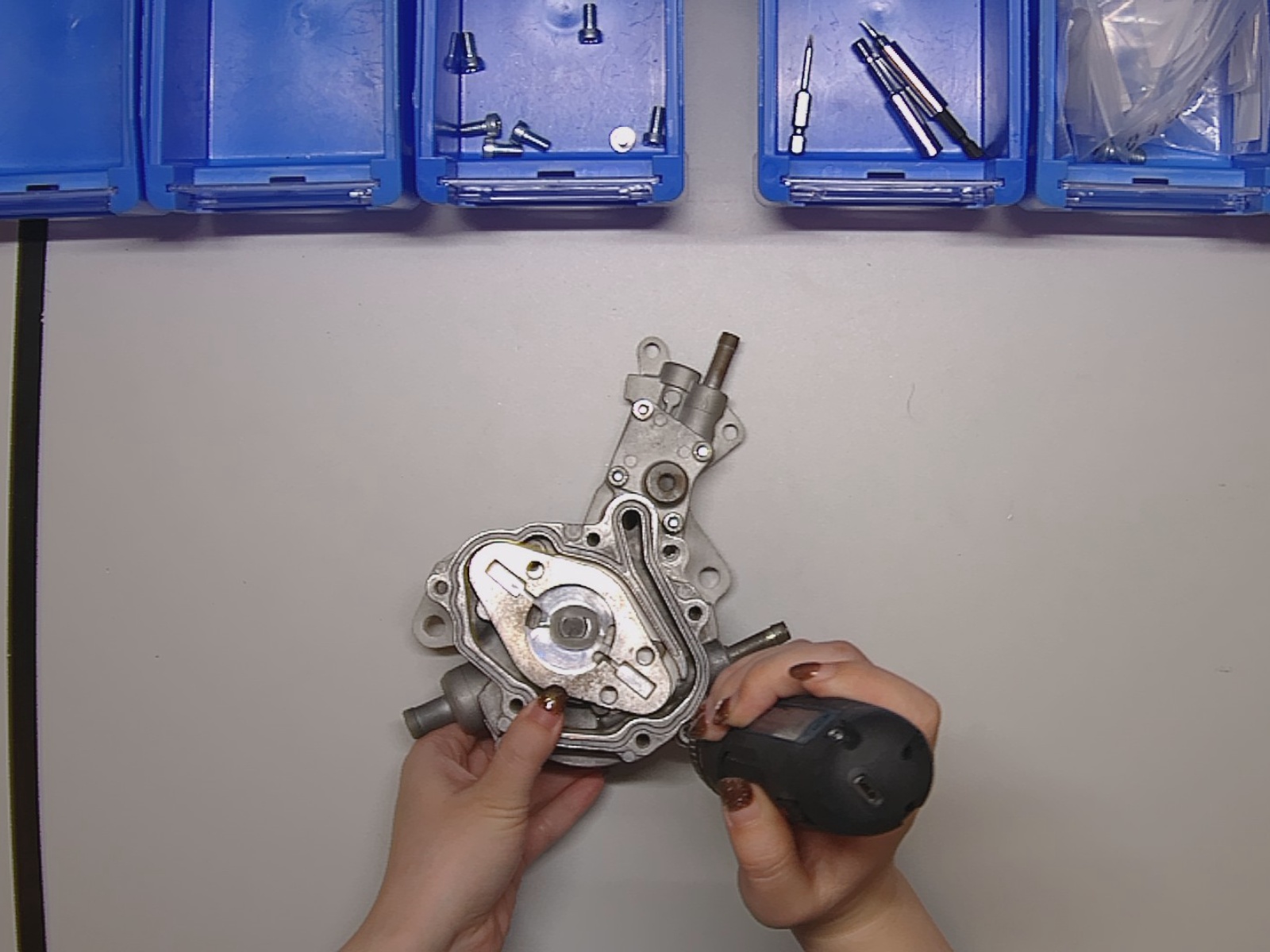}
    \vspace{-1.5mm}
    \caption{Human variations in assembly execution for the same step.}
    \vspace{1.5mm}
    \label{fig:variation_people}
\end{subfigure}
	\begin{subfigure}[b]{0.99\linewidth}
		\centering
		\begin{tabular}{cc}
			\includegraphics[width=0.31\linewidth]{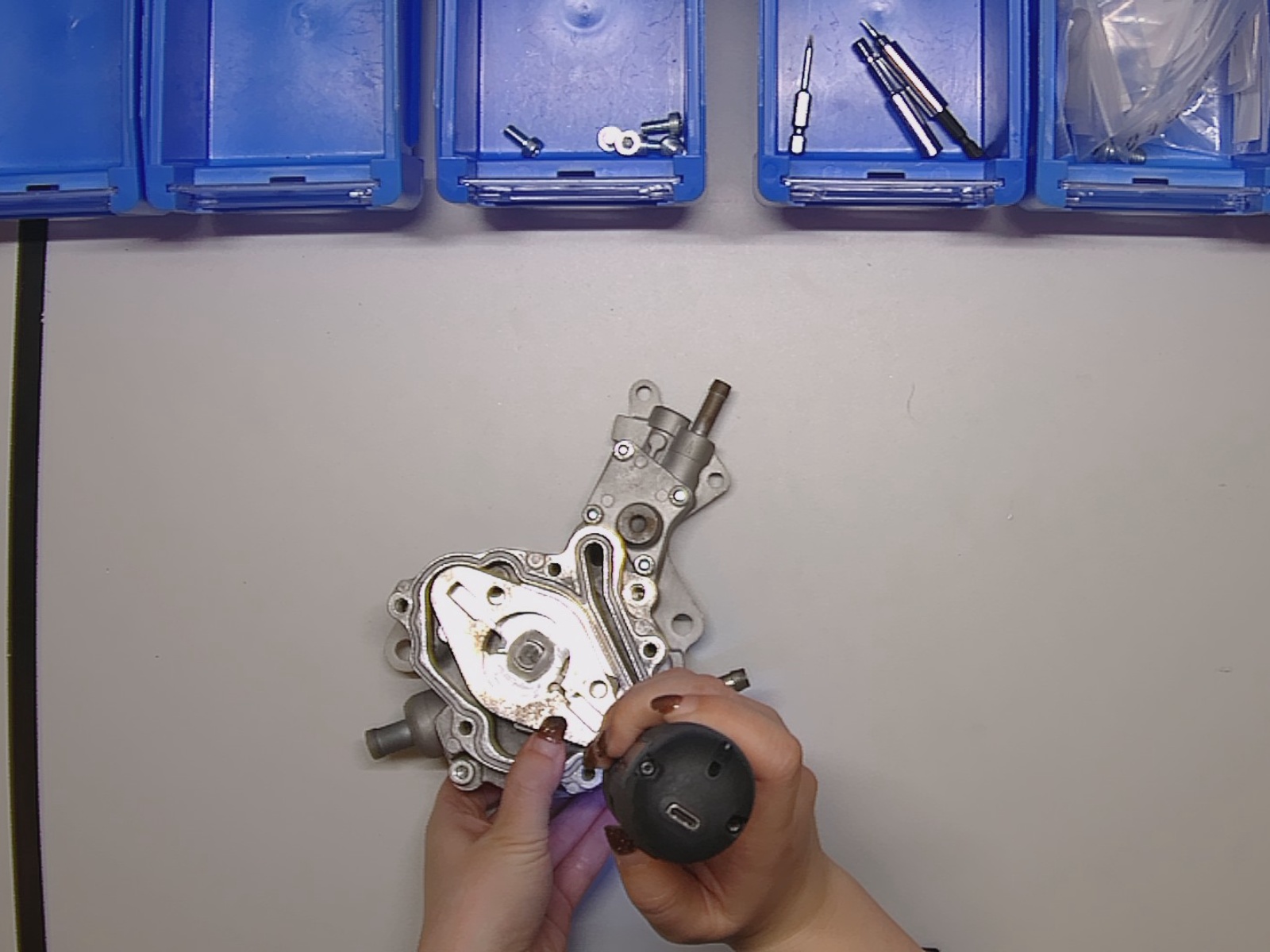} &
			\includegraphics[width=0.31\linewidth]{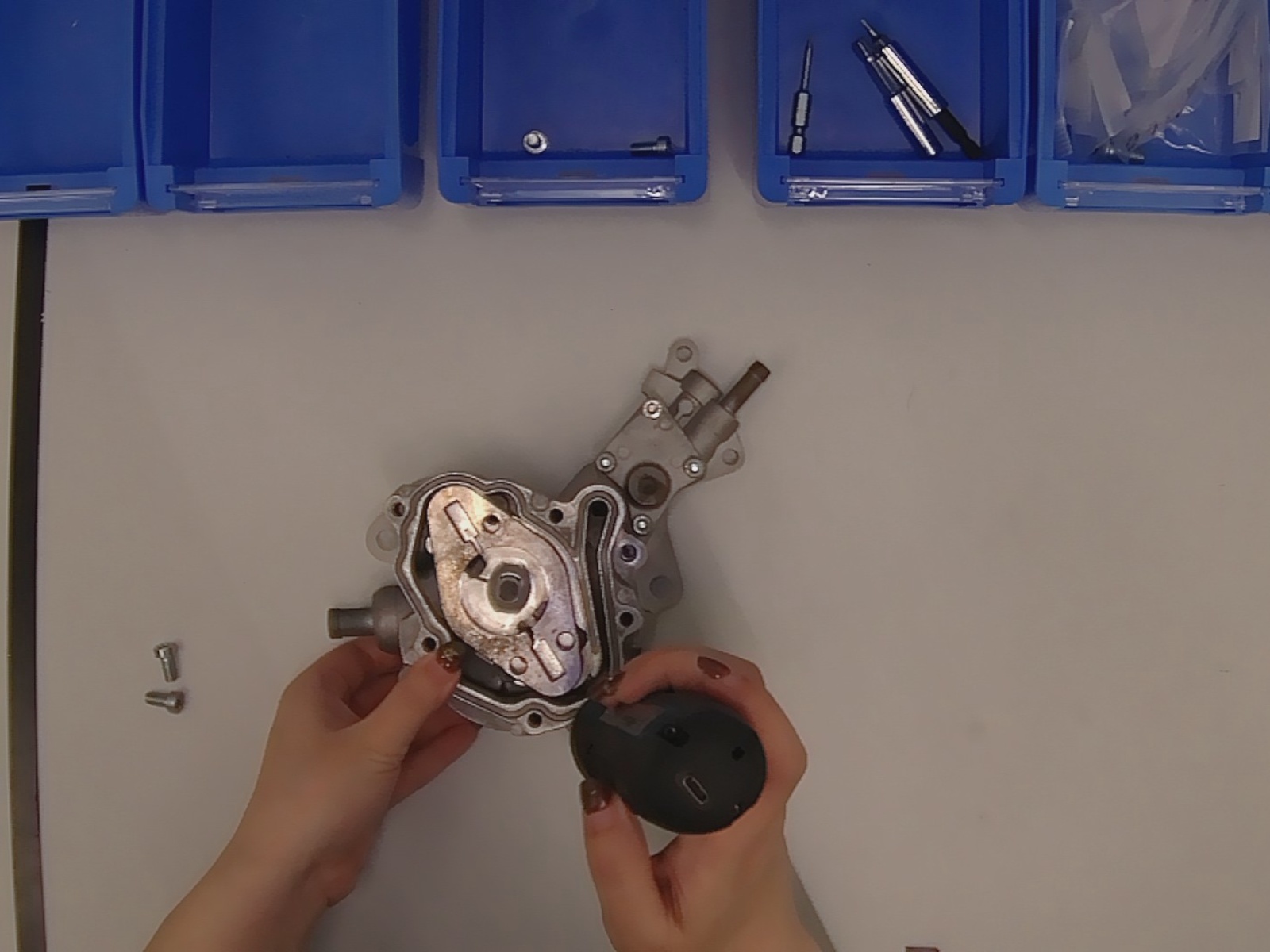} \\
		\end{tabular}
		\vspace{-2.5mm}
		\caption{Lighting variations in assembly execution with the same step.}
		\label{fig:variation_lighting}
	\end{subfigure}
    \vspace{-4.5mm}
	\caption{Visualization of assembly variations: (a) different individuals performing the same assembly step; (b) same individual performing under varying lighting conditions.}
    \vspace{1mm}
	\label{fig:variation}
\end{figure}

Different operators often assemble the same components in distinct ways, as illustrated in Figure~\ref{fig:variation_people}, influenced in particular by factors such as handedness and assembly experience, which poses a common challenge for step recognition. In addition, variations in hand size, shape, and appearance across different ages and genders can further affect the assembly process. To assess the robustness of our method to such human variance, we evaluate its performance across different operators.

\begin{table}[t]
    \centering
\vspace{2mm}
    \resizebox{\columnwidth}{!}{
    \begin{tabular}{l|ccc}
        \toprule
        \textbf{Operator ID} & \textbf{Step Accuracy (\%)} & \textbf{Det. Precision (\%)} & \textbf{Det. Recall (\%)} \\
        \midrule
        Operator 1 & 92.1 & 97.9 & 72.1 \\
        Operator 2 & 96.0 & 98.0 & 77.7 \\
        Operator 3 & 95.0 & 96.4 & 74.9 \\
        Operator 4 & 93.6 & 97.0 & 75.5 \\
        Operator 5 & 88.0 & 97.9 & 75.4 \\
        Operator 6 & 89.7 & 97.3 & 77.5 \\
        \bottomrule
    \end{tabular}
    }
    \caption{Evaluation of robustness to human variance}
    \vspace{-1mm}
    \label{tab:human_variance}
\end{table}

\begin{table}[t]
    \centering
    \resizebox{\columnwidth}{!}{
    \begin{tabular}{l|ccc}
        \toprule
        \textbf{Lighting Condition} & \textbf{Step Accuracy (\%)} & \textbf{Det. Precision (\%)} & \textbf{Det. Recall (\%)} \\
        \midrule
        Normal Lighting & 93.26 & 97.89 & 85.33 \\
        Low Lighting    & 91.52 & 96.86 & 65.65 \\
        \bottomrule
    \end{tabular}
    }
    \caption{Evaluation of robustness to lighting variations}
    \vspace{-5mm}
    \label{tab:lighting_comparison}
\end{table}

The results in Table~\ref{tab:human_variance} demonstrate the robustness of our approach across different human operators. While there is some variation in step recognition accuracy across the six operators (ranging from 88.0\% to 96.0\%), the performance remains consistently high overall, with a mean of 92.4\%. This indicates that our system is resilient to differences in individual assembly behaviors and execution styles. The object detection results are also stable, with precision consistently above 96.4\% and recall ranging narrowly from 72.1\% to 77.7\%, suggesting reliable performance across users. The consistent performance across different operators demonstrates that our diverse synthetic data generation effectively captures human assembly variations, enabling the trained model to learn shared features across different users and generalize well to varying execution styles.

\begin{table}[t]
    \centering
    \vspace{-1mm}
    \resizebox{0.8\columnwidth}{!}{
    \begin{tabular}{ccc}
        \toprule
         \textbf{Step Accuracy (\%)} & \textbf{Det. Precision (\%)} & \textbf{Det. Recall (\%)} \\
        \midrule
         94.7 & 98.9 & 87.7 \\
        \bottomrule
    \end{tabular}
    }
    \caption{Evaluation on the Framework laptop case}
    \vspace{-5mm}
    \label{tab:framework}
\end{table}

\subsubsection{Robustness to Lighting Variations}

Similarly, we also analyze the performance of our model with different lighting conditions. The results are shown in Table~\ref{tab:lighting_comparison}, which reveal the system's robustness to environmental changes. Under normal lighting conditions (with the light on), our method achieves 93.26\% step accuracy with 97.89\% detection precision and 85.33\% recall. While performance slightly decreases under low lighting conditions (with the light off)
the system maintains reasonable effectiveness. The most significant impact is on detection recall, which drops by approximately 20\% under low lighting, due to reduced visibility of metallic components and increased shadows that obscure object boundaries. Nevertheless, the maintained high precision (96.86\%) and reasonable step accuracy (91.52\%) demonstrate that our approach can operate effectively across varying lighting conditions, making it suitable for real industrial environments where lighting may fluctuate.

\subsection{Generalizability to New Cases}
\label{sec:generalizability}
To verify the generalizability of our method, we apply it to a new assembly task, the Framework laptop, which is shown in Figure~\ref{fig:recognition_framework}. This modular laptop that can be assembled in 5 sequential steps: (1) Place the Base; (2) Assemble the Main Keyboard; (3) Assemble the Side Keyboard; (4) Assemble the Touchpad; (5) Assemble the Side Bars. Following the setup of the vacuum pump case, we record 10 assembly sequences from 5 different operators and annotate step labels for evaluation, leading to a total of 7,500 labeled images. With approximately 25 minutes of manual effort (primarily for defining the assembly description JSON file and adjusting component CAD textures for more realistic appearance), our pipeline successfully generates the synthetic data and trains the model, leading to consistently high performance as shown in Table~\ref{tab:framework}. This demonstrates the practical utility of our pipeline without requiring costly real-world data collection and task-specific parameter tuning, enhancing its applicability in real-world industrial environments.

\section{Conclusion and Limitations}
\label{sec:conclusion}

This paper presents a comprehensive system for industrial assembly step recognition that addresses data scarcity through end-to-end synthetic data generation and training. The system integrates three components: physics-based assembly motion generation that simulates human variance with diverse hand–object interactions, photorealistic rendering under randomized conditions to handle visual complexity, and rule-based recognition grounded in object detection to further enhance robustness. 
Evaluation on vacuum pump assembly shows the system, trained solely on synthetic data, achieves 92.4\% accuracy across varying operators and lighting conditions, and can be efficiently adapted to new cases within 25 minutes.
Our system is currently limited by hand–object motion generation, restricting the synthesis of highly agile and complex actions. This constrains the detection of fine-grained pose errors and steps without clear object interactions, which we leave for future work.

\bibliographystyle{IEEEtran}
\bibliography{egbib}

@String(CVPR   = {Proc. IEEE Conf. Comput. Vis. Pattern Recognit. (CVPR)})

@String(ICRA   = {Proc. IEEE Int. Conf. Robot. Autom. (ICRA)})

@String(ICCV   = {Proc. IEEE Int. Conf. Comput. Vis. (ICCV)})

@String(ECCV   = {Proc. Eur. Conf. Comput. Vis. (ECCV)})

@String(RAL    = {IEEE Robot. Autom. Lett.})

@InProceedings{zhang2024graspxl,
title={{GraspXL}: Generating Grasping Motions for Diverse Objects at Scale},
author={Zhang, Hui and Christen, Sammy and Fan, Zicong and Hilliges, Otmar and Song, Jie},
booktitle={European Conference on Computer Vision (ECCV)},
year={2024}
}

@InProceedings{foundationposewen2024,
author        = {Bowen Wen and Wei Yang and Jan Kautz and Stan Birchfield},
title         = {{FoundationPose}: Unified 6D Pose Estimation and Tracking of Novel Objects},
booktitle     = CVPR,
year          = {2024}
}

@InProceedings{wen2025foundationstereo,
  title={{FoundationStereo}: Zero-Shot Stereo Matching},
  author={Bowen Wen and Matthew Trepte and Joseph Aribido and Jan Kautz and Orazio Gallo and Stan Birchfield},
  booktitle=CVPR,
  year={2025}
}

@inProceedings{zhang2024artigrasp,
  title={{ArtiGrasp}: Physically Plausible Synthesis of Bi-Manual Dexterous Grasping and Articulation},
  author={Zhang, Hui and Christen, Sammy and Fan, Zicong and Zheng, Luocheng and Hwangbo, Jemin and Song, Jie and Hilliges, Otmar},
  booktitle={International Conference on 3D Vision (3DV)},
  year={2024}
}

@inproceedings{zhang2025RobustDexGrasp,
      title={{RobustDexGrasp}: Robust Dexterous Grasping of General Objects},
      author={Zhang, Hui and Wu, Zijian and Huang, Linyi and Christen, Sammy and Song, Jie},
      booktitle={Conference on Robot Learning (CoRL)},
      year={2025}
    }

@article{huang2024fungrasp,
  title={{FunGrasp}: Functional Grasping for Diverse Dexterous Hands},
  author={Huang, Linyi and Zhang, Hui and Wu, Zijian and Christen, Sammy and Song, Jie},
  journal=RAL,
  year={2025}
}

@article{zhang2021manipnet,
author = {Zhang, He and Ye, Yuting and Shiratori, Takaaki and Komura, Taku},
title = {{ManipNet}: Neural Manipulation Synthesis with a Hand-Object Spatial Representation},
year = {2021},
journal = {ACM Trans. Graph.},
}

@inproceedings{zheng2023cams,
  title={{CAMS}: CAnonicalized Manipulation Spaces for Category-Level Functional Hand-Object Manipulation Synthesis},
  author={Zheng, Juntian and Zheng, Qingyuan and Fang, Lixing and Liu, Yun and Yi, Li},
  booktitle=CVPR,
  year={2023}
}

@article{Jonas,
author = {Conrad, Jonas and Stauffer, Tobias and Meng, Xuanting and Ferchow, Julian and Meboldt, Mirko},
year = {2024},
title = {Deep learning-based error recognition in manual cable assembly using synthetic training data},
journal = {Procedia CIRP}
}

@article{Zhu2024,
author = {Zhu, Xiaomeng and Martensson, Par and Hanson, Lars and Bjorkman, Marten and Maki, Atsuto},
year = {2024},
title = {Automated assembly quality inspection by deep learning with 2D and 3D synthetic CAD data},
journal = {J. Intell. Manuf.},
}

@article{Mazzetto_2020,
title={Deep Learning Models for Visual Inspection on Automotive Assembling Line},
journal = {Int. J. Adv. Eng. Res. Sci.},
author={Mazzetto, Muriel and Teixeira, Marcelo and Rodrigues, Erick Oliveira and Casanova, Dalcimar},
year={2020}}

@Article{s20154208,
AUTHOR = {Chen, Chengjun and Zhang, Chunlin and Wang, Tiannuo and Li, Dongnian and Guo, Yang and Zhao, Zhengxu and Hong, Jun},
TITLE = {Monitoring of Assembly Process Using Deep Learning Technology},
JOURNAL = {Sensors},
YEAR = {2020},
}

@INPROCEEDINGS{5381107,
  author={Jia, Jiancheng},
  booktitle = {Proc. Int. Conf. Mach. Vis. (ICMV)},
  title={A Machine Vision Application for Industrial Assembly Inspection}, 
  year={2009}}

@INPROCEEDINGS{509207,
  author={Khawaja, K.W. and Maciejewski, A.A. and Tretter, D. and Bouman, C.A.},
  booktitle = {Proc. IEEE Int. Conf. Robot. Autom. (ICRA)},
  title={Camera and light placement for automated assembly inspection}, 
  year={1996}}

@ARTICLE{9383264,
  author={Li, Xuan and Wang, Sukai and Chen, Ping and Wang, Liming},
  journal = {IEEE Trans. Instrum. Meas.},
  title={3-D Inspection Method for Industrial Product Assembly Based on Single X-Ray Projections}, 
  year={2021}}

@inproceedings{shamil2024handformer,
  title={On the Utility of 3D Hand Poses for Action Recognition},
  author={Shamil, Md Salman and Chatterjee, Dibyadip and Sener, Fadime and Ma, Shugao and Yao, Angela},
  booktitle=ECCV,
  year={2024},
}

@inproceedings{zhang2025bimart,
  title={Bimart: A unified approach for the synthesis of 3d bimanual interaction with articulated objects},
  author={Zhang, Wanyue and Dabral, Rishabh and Golyanik, Vladislav and Choutas, Vasileios and Alvarado, Eduardo and Beeler, Thabo and Habermann, Marc and Theobalt, Christian},
  booktitle=CVPR,
  year={2025}
}

@inproceedings{FirstPersonAction_CVPR2018,
  title={First-Person Hand Action Benchmark with RGB-D Videos and 3D Hand Pose Annotations},
  author={Garcia-Hernando, Guillermo and Yuan, Shanxin and Baek, Seungryul and Kim, Tae-Kyun},
  booktitle = CVPR,
  year = {2018}
}

@INPROCEEDINGS{Guilhem2015PCNN,
  author={Chéron, Guilhem and Laptev, Ivan and Schmid, Cordelia},
  booktitle=ICCV, 
  title={P-CNN: Pose-Based CNN Features for Action Recognition}, 
  year={2015}}

@inproceedings{fan2023arctic,
  title = {{ARCTIC}: A Dataset for Dexterous Bimanual Hand-Object Manipulation},
  author = {Fan, Zicong and Taheri, Omid and Tzionas, Dimitrios and Kocabas, Muhammed and Kaufmann, Manuel and Black, Michael J. and Hilliges, Otmar},
  booktitle = CVPR,
  year = {2023}
}

@inproceedings{liu2024taco,
  title={TACO: Benchmarking Generalizable Bimanual Tool-ACtion-Object Understanding},
  author={Liu, Yun and Yang, Haolin and Si, Xu and Liu, Ling and Li, Zipeng and Zhang, Yuxiang and Liu, Yebin and Yi, Li},
  booktitle = CVPR,
  year={2024}
}

@inproceedings{sener2022assembly101,
    title = {Assembly101: A Large-Scale Multi-View Video Dataset for Understanding Procedural Activities},
    author={Fadime Sener and Dibyadip Chatterjee and Daniel Shelepov and Kun He and Dipika Singhania and Robert Wang and Angela Yao},
    booktitle = CVPR,
    year={2022}}

@Manual{blender,
   title = {Blender - a 3D modelling and rendering package},
   author = {{Blender Online Community}},
   url = {http://www.blender.org},
}

@software{yolov8_ultralytics,
  author = {Glenn Jocher and Ayush Chaurasia and Jing Qiu},
  title = {Ultralytics YOLOv8},
  version = {8.0.0},
  year = {2023},
  url = {https://github.com/ultralytics/ultralytics},
  orcid = {0000-0001-5950-6979, 0000-0002-7603-6750, 0000-0003-3783-7069},
  license = {AGPL-3.0}
}

@article{MANO:SIGGRAPHASIA:2017,
    title = {Embodied Hands: Modeling and Capturing Hands and Bodies Together},
    author = {Romero, Javier and Tzionas, Dimitrios and Black, Michael J.},
    journal = {ACM Transactions on Graphics, (Proc. SIGGRAPH Asia)},
    publisher = {ACM},
    year = {2017},
}

@ARTICLE{Hwangbo2018Raisim,
  author={Hwangbo, Jemin and Lee, Joonho and Hutter, Marco},
  journal=RAL, 
  title={Per-Contact Iteration Method for Solving Contact Dynamics}, 
  year={2018},
}

@inproceedings{fu2025gigahands,
  title={GigaHands: A Massive Annotated Dataset of Bimanual Hand Activities},
  author={Fu, Rao and Zhang, Dingxi and Jiang, Alex and Fu, Wanjia and Fund, Austin and Ritchie, Daniel and Sridhar, Srinath},
  year={2025},
  booktitle=CVPR
}

@inproceedings{wang2023pov,
  title={Pov-surgery: A dataset for egocentric hand and tool pose estimation during surgical activities},
  author={Wang, Rui and Ktistakis, Sophokles and Zhang, Siwei and Meboldt, Mirko and Lohmeyer, Quentin},
  booktitle = {Proc. Int. Conf. Med. Image Comput. Comput.-Assist. Interv. (MICCAI)},
  year={2023},
}

@inproceedings{zhan2024oakink2,
      title={OAKINK2: A Dataset of Bimanual Hands-Object Manipulation in Complex Task Completion}, 
      author={Xinyu Zhan and Lixin Yang and Yifei Zhao and Kangrui Mao and Hanlin Xu and Zenan Lin and Kailin Li and Cewu Lu},
      year={2024},
      booktitle=CVPR
}

@misc{rawal2024syntheticdatagenerationbridging,
      title={Synthetic Data Generation for Bridging Sim2Real Gap in a Production Environment}, 
      author={Parth Rawal and Mrunal Sompura and Wolfgang Hintze},
      year={2024},
      eprint={2311.11039},
      archivePrefix={arXiv},
      primaryClass={cs.CV},
}

@inproceedings{schieber2024asdf,
  title={Asdf: Assembly state detection utilizing late fusion by integrating 6d pose estimation},
  author={Schieber, Hannah and Li, Shiyu and Corell, Niklas and Beckerle, Philipp and Kreimeier, Julian and Roth, Daniel},
  booktitle={2024 IEEE International Symposium on Mixed and Augmented Reality (ISMAR)},
  pages={190--199},
  year={2024},
  organization={IEEE}
}

@article{wang2025learning,
  title={Learning Generalizable Hand-Object Tracking from Synthetic Demonstrations},
  author={Wang, Yinhuai and Yu, Runyi and Tsui, Hok Wai and Lin, Xiaoyi and Zhang, Hui and Zhao, Qihan and Fan, Ke and Li, Miao and Song, Jie and Wang, Jingbo and others},
  journal={arXiv preprint arXiv:2512.19583},
  year={2025}
}

@article{smirl,
title = {Real-time action localization of manual assembly operations using deep learning and augmented inference state machines},
journal = {Journal of Manufacturing Systems},
year = {2024},
author = {Vignesh Selvaraj and Md Al-Amin and Xuyong Yu and Wenjin Tao and Sangkee Min},
}

@inproceedings{aganian2023attach,
  title={ATTACH Dataset: Annotated Two-Handed Assembly Actions for Human Action Understanding},
  author={Aganian, Dustin and Stephan, Benedict and Eisenbach, Markus and Stretz, Corinna and Gross, Horst-Michael},
  booktitle=ICRA,
  year={2023},
}

\end{document}